\documentclass[journal,onecolumn,letterpaper]{IEEEtran}

\usepackage[utf8]{inputenc}
\usepackage{amsmath,amsfonts}
\usepackage{algorithmic}
\usepackage{algorithm}
\usepackage{array}
\usepackage[caption=false,font=normalsize,labelfont=sf,textfont=sf]{subfig}

\usepackage{textcomp}
\usepackage{stfloats}
\usepackage{url}
\usepackage{verbatim}
\usepackage{graphicx}
\usepackage{cite}
\usepackage{multicol}
\usepackage{subfig}
\usepackage{xcolor}
\usepackage{pdflscape}
\usepackage{mathtools, cuted}
\usepackage{csquotes}
\usepackage{ulem}
\usepackage{booktabs} 
\usepackage{adjustbox} 
\usepackage{multicol}
\usepackage{multirow}
\usepackage{enumitem}
\usepackage{algorithm}
\usepackage{algorithmic}

\setlist[itemize]{
    leftmargin=*,  
    itemsep=0pt,   
    topsep=0pt,
    parsep=0pt
}
\setlist[enumerate]{
    leftmargin=*,  
    itemsep=0pt,   
    topsep=0pt,
    parsep=0pt
}
\begin{document}
\title{On the Equivalence of Regression and Classification}
\author{Jayadeva$^{*1}$, Naman Dwivedi$^1$, Hari Krishnan C. K.$^1$,  N. M. Anoop Krishnan$^2$\\
 $^1$ Department of Electrical Engineering, $^2$Department of Civil Engineering\\
       Indian Institute of Technology Delhi, Hauz Khas, New Delhi 110016}



\maketitle

\begin{abstract}
A formal link between regression and classification has been tenuous. Even though the margin maximization term $\|w\|$ is used in support vector regression, it has at best been justified as a regularizer. We show that a regression problem with $M$ samples lying on a hyperplane has a one-to-one equivalence with a linearly separable classification task with $2M$ samples. We show that margin maximization on the equivalent classification task leads to a different regression formulation than traditionally used. Using the equivalence, we demonstrate a ``regressability'' measure, that can be used to estimate the difficulty of regressing a dataset, without needing to first learn a model for it. We use the equivalence to train neural networks to learn a linearizing map, that transforms input variables into a space where a linear regressor is adequate.
\end{abstract}

\textbf{Keywords}: Regression, Computation Theory, Optimal Map\\

\section{Introduction}
Both classification and regression formulations typically include $L_1$ or $L_2$ norms of the weight vector, i.e. $\|w\|_p, \; p = 1, 2$ in their optimization objectives. The primal Support Vector Classifier (SVC) classifier minimizes the $\mathcal{L_2}$ norm of the weight vector, \textit{viz.}  $\frac{1}{2}\|w\|^2$, in order to maximize the margin \textit{viz.} $\frac{2}{\|w\|}$. However, there is no equivalent notion of margin in the case of support vector regression (SVR), and the use of the same term in regression formulations is at best justified as a regularization term.

 Torgo and Gama \cite{torgo1996regression} split the regressed output range into a set of levels and assigned the regression task to a classification problem with multiple classes. A somewhat similar strategy was adopted by \cite{salman2012regression}. Equivalence in the context of SVC and $\epsilon$-SVR was shown by \cite{pontil1998regression}. They assume that "as in the classification case, the objective is to find a tradeoff between
finding a hyperplane with the small norm and finding a hyperplane that performs regression well". The margin of the regressing hyperplane is $\frac{\|w\|}{1-\epsilon}$, where $w$ is the weight vector of the hyperplane.\\

Bi and Bennett $[6]$ attempted to show that in the context of $\epsilon$-SVR, every regression task with $M$ samples $\left( x^i, z_i \right)$, $x^i \in \Re^n$, $z_i \in \Re$, $i$ = 1, 2, ..., $M$ can be mapped to an equivalent SVC with $2M$ samples. In the equivalent SVC task, $M$ Class 1 samples are located at $\left( x^i, z_i + \epsilon \right)$, and another $M$ Class (-1) samples are located at $\left( x^i, z_i - \epsilon \right)$. 

\begin{figure}[h]
\centering
\includegraphics[width=0.4\textwidth]{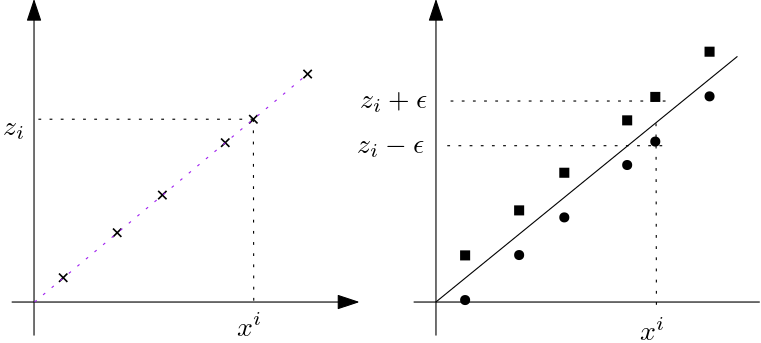}
  \caption{Left: $M$ regression samples on a line. Right: samples for the classification task obtained by duplicating and shifting samples by $\pm \epsilon$}
  \label{bibennett1}
\end{figure}

A hyperplane $w^Tx +b = 0$ that solves the $\epsilon$-SVR must satisfy
\begin{gather}
    w^Tx + b \geq z_i -\epsilon \nonumber\\
    w^Tx + b \leq z_i +\epsilon \nonumber.
\end{gather}
Bi \& Bennett show that in the equivalent SVC, a hyperplane boundary of the form
\begin{gather}
    w^Tx + b + \eta z = 0\nonumber
\end{gather}
must pass between Class 1 samples $\{x^i, z_i + \epsilon\}$, and Class -1 samples $\{x^i, z_i - \epsilon\}$. Figure \ref{bibennett1} illustrates their approach for an example in which all samples to be regressed lie on a straight line.

Once such a hyperplane is learned by solving the classification problem, the regression estimate of an unknown sample $x$ may be found from
\begin{gather}
    z = \frac{-(w^Tx + b)}{\eta}
\end{gather}
However, note that with this equivalence, the margin is $2 \epsilon$, independent of $w$, since the separating hyperplane must pass in between samples of the two classes.

The organization of the remainder of the paper, and key contributions, are summarized below.\\

Consider a regression dataset $\{x^i, z_i\},~ i=$ 1, 2 ..., $M$, where $x^i \in \Re^n$, $z_i \in \Re$. We first consider the case when all samples lie exactly on a hyperplane. \\
$\bullet \; $ We show that the $M$ class 1 samples located at $\frac{x^i}{z_ i}$ and $M$ class "-1" samples located at $-\frac{x^i}{z_i}$ are support vectors of an equivalent linearly separable binary classification problem.\\
This equivalence between regression and classification is discussed in Section \ref{equivalence}.\\
$\bullet \; $ Section \ref{regressability} discusses the proposed \textit{regressability} measure, that estimates the difficulty of regressing a dataset without needing to first learn a regressor. The measure uses the equivalence shown in Section \ref{equivalence}.\\
$\bullet \; $ The equivalence allows us to train a neural network to learn a linearizing map $\phi(x)$. This is a map that transforms input variables $x$ into a higher dimensional feature space $\phi(x)$, such that the regressed output $z$ is a linear function of $\phi(x)$, i.e. $z = w^T\phi(x)$, where $w$ is a weight or co-efficient vector. This is discussed in Section \ref{J4}.\\
$\bullet \; $ We learn a regressor in 2 steps. First, we learn a linearizing map as mentioned; we then learn a linear hyperplane in the feature space $\phi(x)$. This two step process obviates the need to tune hyperparameters that control the tradeoff between different terms of a loss function.
$\bullet \; $ Learning a linear regressor on such a feature map provides smoother interpolation, and in some cases, provides some extrapolation ability as well.\\
$\bullet \; $ Section \ref{results} discusses results and comparisons. Section \ref{conclusion} contains concluding remarks.

\section{Regression is Classification}\label{equivalence}
\section{Approach to convert regression to equivalent classification problem}
We first consider a SVC problem with a linerly separable set of samples $\{\hat{x}^i, y_i\},~ i=$ 1, 2 ..., $M$, where $\hat{x}^i \in \Re^n$, $y_i \in \{-1, 1\}$. We also assume that the separating hyperplane $w^Tx = 0$ passes through the origin; we revisit this assumption in the sequel. The primal hard margin support vector machine classifier solves
\begin{gather}
    \text{Minimize}_{~w} ~ \frac{1}{2} \|w\|^2 \\
\text{subject to the constraints} \nonumber\\
y_i \left(w^Tx^i \right) \geq 1,\; i = 1, 2, ..., M
\end{gather}
In this case, all support vectors satisfy
\begin{gather}
y_i \left(w^Tx^i \right) = 1, \text{i.e.}\\
\left(w^Tx^i \right) = 1,~ \text{for class 1 samples, and} \label{cl1}\\
-\left(w^Tx^i \right) = -1,~ \text{for class -1 samples, and} \label{cl-1}
\end{gather}
Now consider a regression dataset $\{x^i, z_i\},~ i=$ 1, 2 ..., $M$, where $x^i \in \Re^n$, $z_i \in \Re \setminus 0$. Without loss of generality, we assume that the desired regressor passes through the origin. We first consider the case when all samples lie on a hyperplane. We have
\begin{gather}
\left(w^Tx^i \right) = z_i.\\
\text{This allows two possibilities}\nonumber\\
w^T\left( \frac{x^i}{z_i} \right) = 1 \label{rcl1}\\
w^T\left( -\frac{x^i}{z_i} \right) = -1 \label{rcl-1}\\
\end{gather}
Constraints (\ref{rcl1}) and (\ref{rcl-1}) are identical to (\ref{cl1}) and (\ref{cl-1}). The equivalence for the "hard margin" case is thus trivially established. In a nutshell, a regression dataset $\{x^i, z_i\},~ i=$ 1, 2 ..., $M$, where $x^i \in \Re^n$, $z_i \in \Re \setminus 0$ is equivalent to a classification task with samples
\begin{gather}
    \left( \frac{x^i}{z_i} \right) \text{ in class 1, i.e. } y_i = 1\text{ , and}\\
    \left( \frac{-x^i}{z_i} \right) \text{ in class -1, i.e. } y_i = -1
\end{gather}
It is also evident that if the regression samples lie on a hyperplane, samples of the equivalent classification task are linearly separable. Let the solution to the classification problem be a hyperplane $w$. Clearly, for all support vectors, $w^Tx = \pm 1$. For the solution to the regression task, the same $w$ suffices, since $w^Tx = z$.\\
We now elucidate the approach with an intuitive example. Consider samples drawn randomly from the line $z = mx~, m \in \Re$; any set of samples will have $z_i = m x_i$. The equivalent classification task will have class -1 support vectors at $\frac{-1}{m}$, and class 1 support vectors at $\frac{+1}{m}$, i.e. there will be only 2 support vectors at $\frac{-1}{m}$ and $\frac{+1}{m}$. The separating hyperplane in the equivalent classification problem is the line
\begin{gather}
    m x = 0
\end{gather}.
The regression estimate $z^*$ for a test sample $x^*$ is given by $z^* = m x^*$. In the equivalent classification problem, the samples corresponding to $x^*$ are given by
\begin{gather}
    \left( \frac{x^*}{z^*} \right) \text{ in class 1, and}\\
    \left( \frac{-x^*}{z^*} \right) \text{ in class -1}
\end{gather}

\begin{figure}[h!]
\centering
\includegraphics[width=0.4\textwidth]{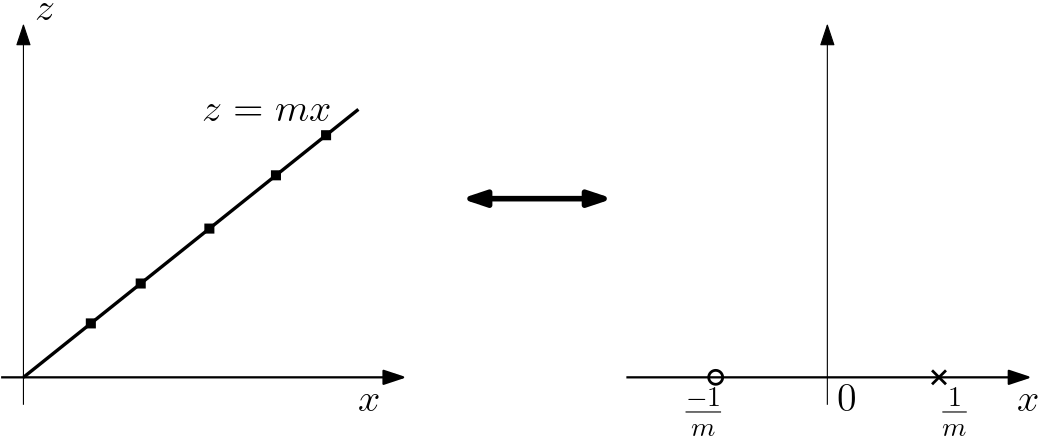}
  \caption{Left: Regression samples taken from the line $z = mx$. Right: Samples for the equivalent classification task lie at $\pm \frac{1}{m}$. Note that the margin for the classification task is $\frac{2}{m}$, that tends to zero as the line becomes more vertical.\\}\label{equiv1d}
\end{figure}

The regression problem and the classification task are shown in Fig. \ref{equiv1d}. Note that the margin in this case is $\frac{2}{m}$.\\

Observe that as the slope of the line $m$ increases, the margin reduces. Any small perturbation of the sample location $x$ to $x + \epsilon$ will lead to an error in the regression estimate by $m(x + \epsilon)$, i.e. the error will increase as the line becomes more vertical. When $m \rightarrow \infty$, $z$ cannot be estimated from $x$, and the margin of the equivalent classification task $\rightarrow 0$.\\

In real world tasks, the permissible regression error is relative to the regressed value, e.g. when plotting the current-voltage (I-V) relationship curve for a diode, the currents may range from microamperes $10^{-6}$ for small values of the voltage V across the diode, to amperes for large values of V. An error of a $\mu$A is negligible when $I = 1A$, but clearly unacceptable for $I = 1 \mu A$. Our approach to regression shows that the equivalent classification task normalizes samples with respect to their location on the regressing hyperplane.\\
Consider a regression task with samples $\{x^i, z_i\},~ i=$ 1, 2 ..., $M$, where $x^i \in \Re^n$, $z_i \in \Re \setminus 0$.
\begin{gather}
    x^i_+ = \left( \frac{x^i}{z_i} \right) \text{ in class 1, i.e. } y_i = 1\text{ , and}\\
    x^i_- = \left( \frac{-x^i}{z_i} \right) \text{ in class -1, i.e. } y_i = -1
\end{gather}\label{equivclass1}
We now show the SVC formulation and a formal equivalence. The SVC formulation is in the spirit of a least squares SVM \cite{suykens1999least} or proximal SVM \cite{fung2001proximal}, but employs a L1 error measure. Consider the SVC formulation where we assume that the separating hyperplane passes through the origin. The classification dataset is given by $\{x^i, y_i\},~ i=$ 1, 2 ..., $M$, where $x^i \in \Re^n$, $y_i \in \{-1, 1\}$
 \begin{gather}
     \text{Minimize } \frac{1}{2}\|w\|^2 + C \sum_{i = 1}^M \left( q_i^+ + q_i^-\right) \label{svcobj}\\`
     \text{subject to the constraints} \nonumber\\
     y_i(w^Tx^i) + \left( q_i^+ - q_i^-\right) = 1 \label{svccon1}\\
     q_i^+, q_i^- \geq 0~, \; i = 1, 2, ..., M \label{svccon2}\\
 \end{gather}

\begin{figure}[h]
\centering
\includegraphics[width=0.4\textwidth]{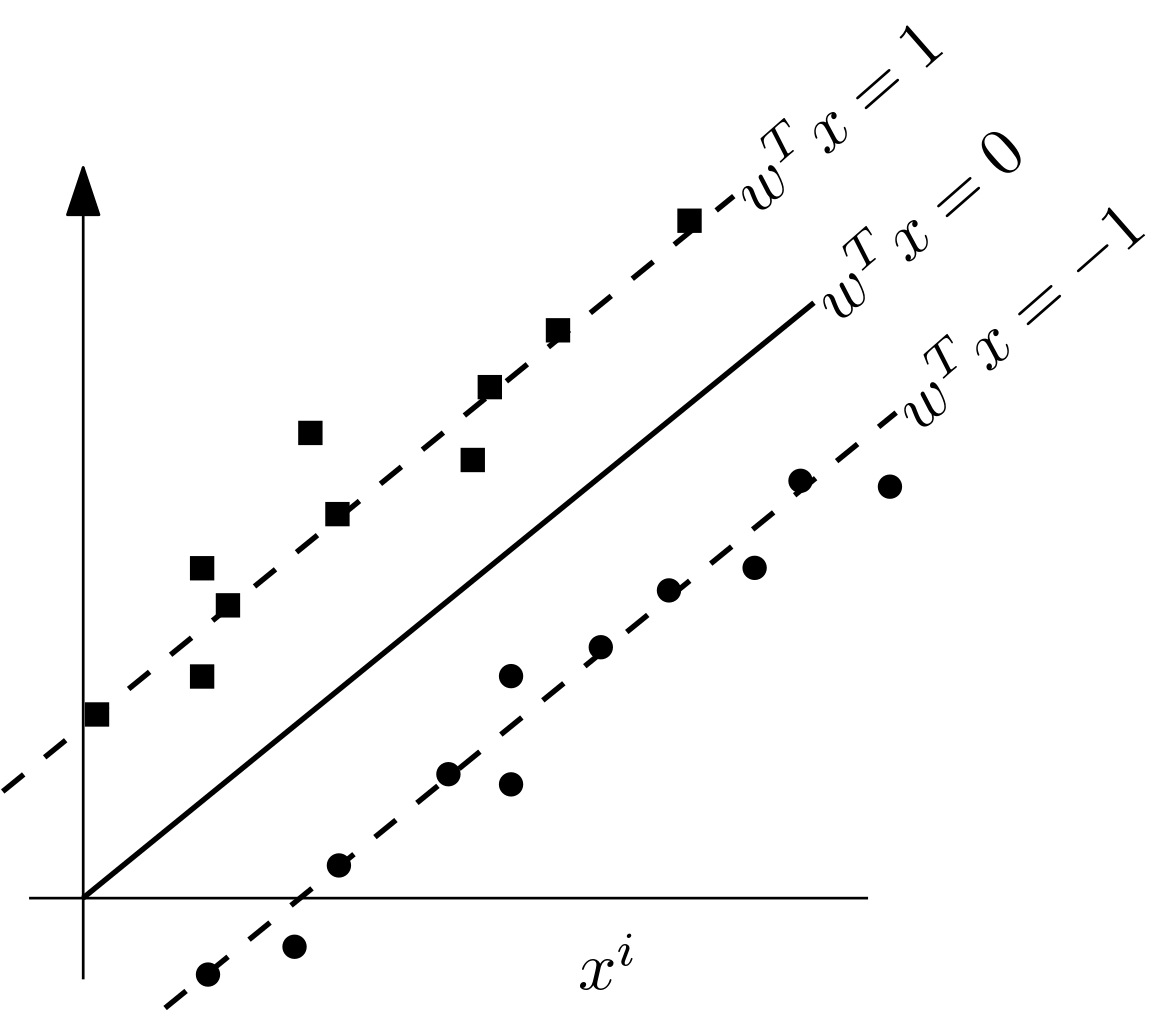}
  \caption{The separating hyperplane is given by $w^Tx = 0$. Hyperplanes $w^Tx = 1$ and $w^Tx = -1$ are proximal to samples of the two classes. A sample $x^i$ not lying on either of these hyperplanes will be at a non-zero distance from its proximal plane, given by $w^Tx^i = z_i - (q_i^+ - q_i^-)$}
  \label{figsvclinear}
\end{figure}

The separating hyperplane is given by $w^Tx = 0$; hyperplanes $w^Tx = 1$ and $w^Tx = -1$ are proximal to the two classes, as in the case of the least squares SVM \cite{suykens1999least} or proximal SVM \cite{fung2001proximal}.  The separating and proximal planes are illustrated in Fig. \ref{figsvclinear}. 

The dual is derived in \ref{Appendix1} and reproduced here for convenience. 
 \begin{gather}
     \text{Minimize } \frac{1}{2}\sum_{i=1}^M\sum_{i=1}^M ~ \lambda_i \lambda_j y_i y_j \left( x^i\right)^T\left( x^j\right) - \sum_{i=1}^M ~ \lambda_i \label{svcdualobj}\\
     \text{subject to the constraints} \nonumber\\
     -C \leq \lambda_i \leq C \label{svccons1}\\
 \end{gather}
  From the K.K.T. conditions (see Appendix ref{Appendix1}), we note that
 \begin{gather}
     -C < \lambda_i < C \implies q_i^+ = q_i^- = 0 \implies y_i(w^Tx^i) = 1\\
     \lambda_i = -C \implies q_i^- = 0 \implies y_i(w^Tx^i) < 1\\
     \lambda_i = C \implies q_i^+ = 0 \implies y_i(w^Tx^i) > 1\\
 \end{gather}
The equivalent classification samples
\begin{gather}
    x^i_+ = \left( \frac{x^i}{z_i} \right) \text{ in class 1, i.e. } y_i = 1\text{ , and}\\
    x^i_- = \left( \frac{-x^i}{z_i} \right) \text{ in class -1, i.e. } y_i = -1
\end{gather}
are then used to solve the SVC (\ref{svcdualobj})-(\ref{svccons1}), which now has $2M$ samples. For convenience, we re-index the samples from $-M$ to $M$ (excluding 0), so that $x^i = x^i_+, i = 1, 2, ..., M$ and $x^i = x^i_-~, \;i = -1, 2, ..., -M$. The Lagrange multipliers $\lambda_i~, i =\pm 1, \pm 2$, $..., \pm M$, are also correspondingly re-indexed.
Note that
\begin{gather}
     -C < \lambda_i < C \implies w^T\left( \frac{x^i}{z_i} \right) = 1 \implies w^Tx^i = z_i ~, \;i = 1, 2, ..., M\\
     \implies w^T\left( \frac{-x^i}{z_i} \right) = -1~, \implies -C < \lambda_i < C,\; i = -1, -2, ..., -M \\
     \lambda_i = -C \implies q_i^- = 0 \implies
     w^T\left( \frac{x^i}{z_i} \right) < 1 \implies w^Tx^i < z_i~, \;i = 1, 2, ..., M\\
     \implies w^T\left( \frac{-x^i}{z_i} \right) > -1 \implies \lambda_i = -C, q_i^+ = 0\; i = -1, -2, ..., -M \\
     \lambda_i = C \implies q_i^+ = 0 \implies
     w^T\left( \frac{x^i}{z_i} \right) > 1 \implies w^Tx^i > z_i~, \;i = 1, 2, ..., M\\
     \implies w^T\left( \frac{-x^i}{z_i} \right) < -1 \implies \lambda_i = C, q_i^+ = 0\; i = -1, -2, ..., -M \\
 \end{gather}
In short, samples $\left(x^i, z_i\right)$ that satisfy on $w^Tx^i = z_i$ correspond to Lagrange multipliers that satisfy $-C \leq \lambda_i \leq C$; those lying above the regressor, i.e. $w^Tx^i  < z_i$ correspond to $\lambda_i = -C$, and those lying below the regressor, i.e. $w^Tx^i  > z_i$ correspond to $\lambda_i = C$.\\

Figure 4 shows regression samples drawn randomly from four known functions of one variable, and equivalent samples for corresponding classification tasks, obtained by using (\ref{equivclass1}). Figure 4(a) shows samples from the line $z = 2x$. In this case, the equivalent classification task has only 2 samples, one at $x = -\frac{1}{2}$, and the other at $x = \frac{1}{2}$, as shown in Fig. 4(e) . The pairs {(b), (f)}, {(c), (g)}, and {(d), (h)} illustrate other functions.

\begin{figure}[!t]
  \centering

  \subfloat[]{%
    \includegraphics[width=0.20\linewidth]{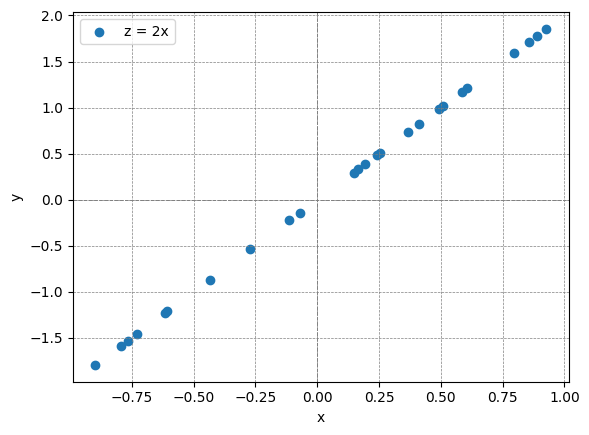}}%
  \hfil
  \subfloat[]{%
    \includegraphics[width=0.20\linewidth]{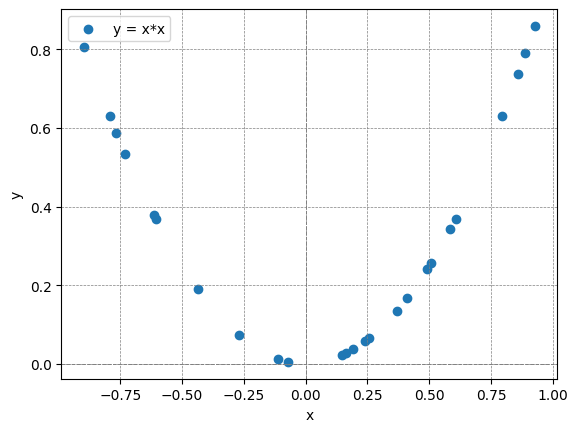}}%
  \hfil
  \subfloat[]{%
    \includegraphics[width=0.20\linewidth]{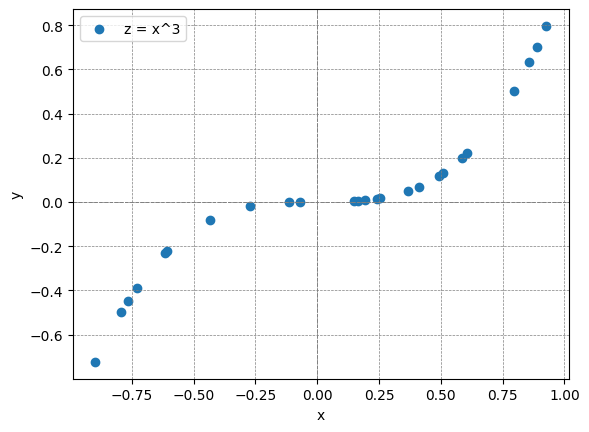}}%
  \hfil
  \subfloat[]{%
    \includegraphics[width=0.20\linewidth]{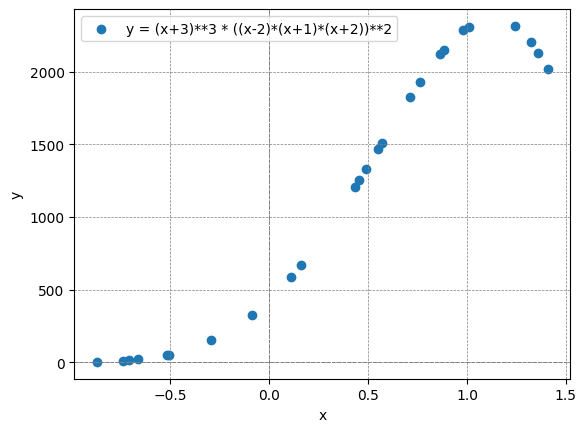}}%

  \par\medskip

  \subfloat[]{%
    \includegraphics[width=0.20\linewidth]{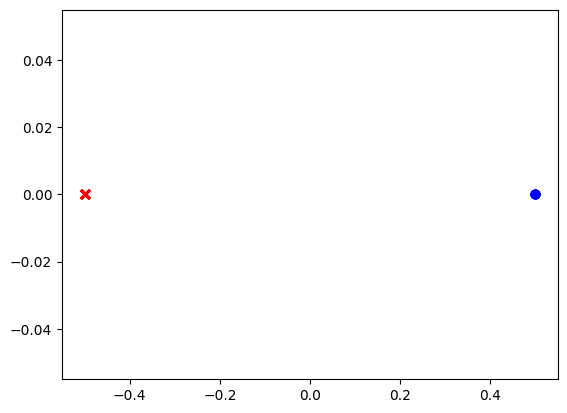}}%
  \hfil
  \subfloat[]{%
    \includegraphics[width=0.20\linewidth]{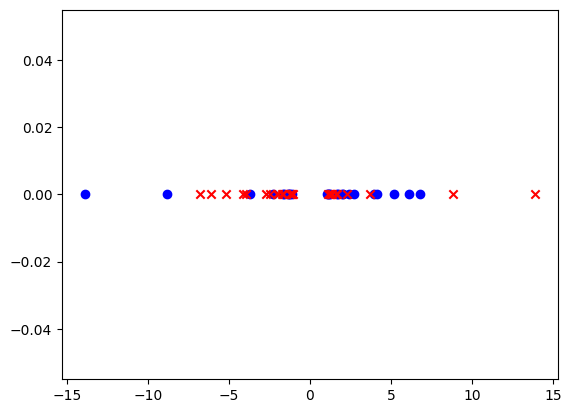}}%
  \hfil
  \subfloat[]{%
    \includegraphics[width=0.20\linewidth]{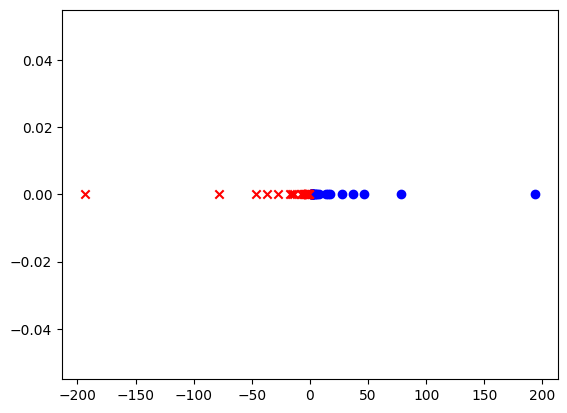}}%
  \hfil
  \subfloat[]{%
    \includegraphics[width=0.20\linewidth]{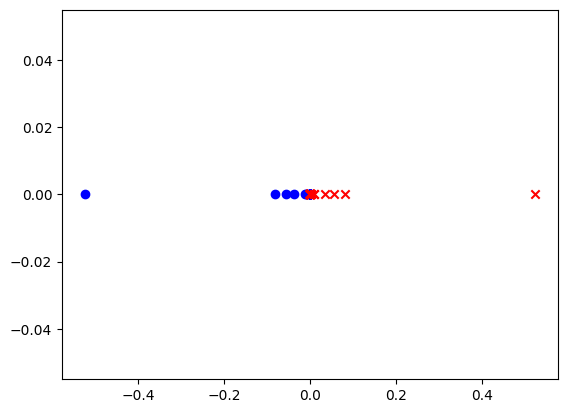}}%

  \caption{(a) $z=2x$, (b) $z=x^2$, (c) $z=x^3$, (d) $z=((x+3)^3)((x-2)(x+1)(x+2))^2$; (e)--(h) equivalent binary classification problems corresponding to (a)--(d), with samples of the two classes distinguished by colour.}
  \label{fig:equiv1d}
\end{figure}

It is evident that other than the case when samples are drawn from a straight line, other functions lead to non-linearly separable classification problems. Dong and Kothari \cite{dong2003feature} proposed a novel approach to estimating the relative difficulty of classifying a labelled set of samples, that they termed as "classifiability". It is not a wrapper approach, and does not require a classifier to be learned, in order to compute the measure. In Section \ref{regressability}, we briefly discuss classifiability, and using our equivalence, propose a "regressability" measure that estimates the difficulty of regressing a set of samples. Illustrative examples highlighted in the section provide insight into why some datasets may be much harder than others.

\section{How difficult is learning a regression dataset : Regressability}\label{regressability}
Figure 5(a)shows a classification problem in two variables; Fig. 5(b) shows the same data with the third dimension being the class label. Figure 6(a)and (b) show the same for another problem, where the data is better separated by a smoother decision boundary. 

\begin{figure}[!t]
  \centering

  \subfloat[]{%
    \includegraphics[width=0.40\linewidth]{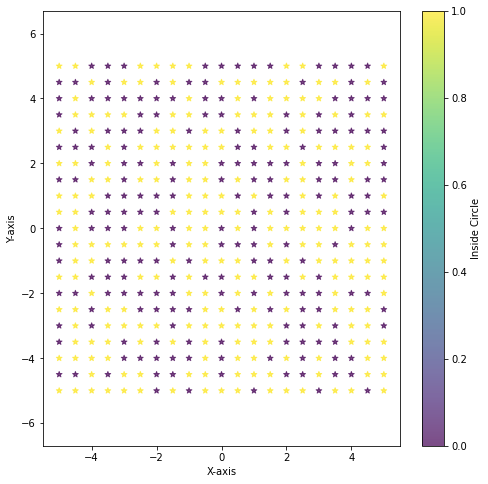}}%
  \hfill
  \subfloat[]{%
    \includegraphics[width=0.40\linewidth]{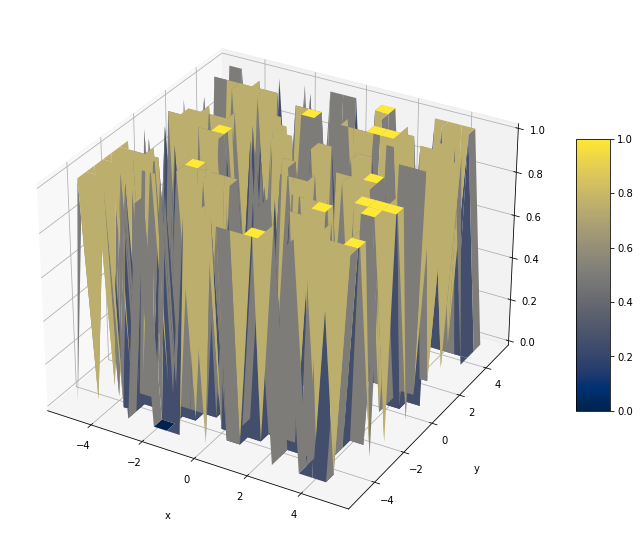}}%

  \caption{(a) Samples of a classification problem and (b) the same samples with the class label indicated by the surface height.}
  \label{fig:c1}
\end{figure}


\begin{figure}[!t]
  \centering

  \subfloat[]{%
    \includegraphics[width=0.40\linewidth]{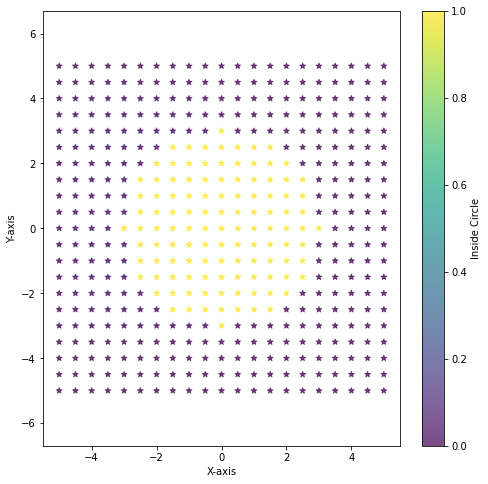}}%
  \hfill
  \subfloat[]{%
    \includegraphics[width=0.40\linewidth]{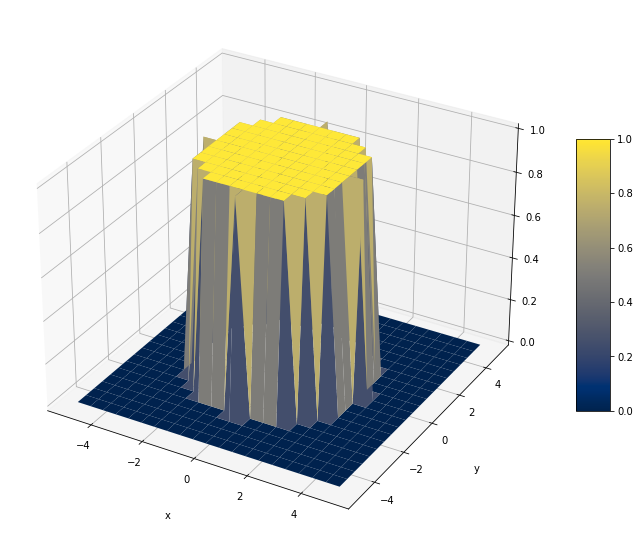}}%

  \caption{(a) Samples of a simple classification problem and (b) the same samples with the surface height equal to the class label.}
  \label{fig:c2}
\end{figure}

Dong and Kothari \cite{dong2003feature} argued that the roughness of the class label surface is a measure of the difficulty of classifying the data, since the surface will be smooth when neighbours with similar labels are adjacent to each other.

Dong and Kothari \cite{dong2003feature} examined the  second order joint conditional density function $f(\omega_1,\omega_2|d)$, i.e. the probability of going from class $\omega_1$ to class $\omega_2$ within a distance $d$. Formally, in a binary classification problem, consider a training sample $x^i$, and another sample $x^j$ lying in its neighbourhood, say within a distance $d$. Assuming that $x^i$ and $x^j$ are independent, the joint probability matrix
\begin{gather}
J^i=\left[\begin{array}{ll}
P\left(\omega_1\left|x^j, \omega_1\right| x^i\right) & P\left(\omega_2\left|x^j, \omega_1\right| x^i\right) \\
P\left(\omega_1\left|x^j, \omega_2\right| x^i\right) & P\left(\omega_2\left|x^j, \omega_2\right| x^i\right)
\end{array}\right]
\end{gather}
will be strongly diagonal when patterns in the neighborhood of $x^i$ belong to $x^i$'s class. As the class label surface becomes more rough, the off-diagonal entries become larger. Classifiability in the neighbourhood of $x^i$ is computed as
\begin{equation}
\begin{aligned}
C\left( x^i \right)= & P\left(\omega_1 \mid x^j\right) P\left(\omega_1 \mid x^i\right) \\
& +P\left(\omega_2 \mid x^j\right) P\left(\omega_2 \mid x^i\right)-P\left(\omega_2 \mid x^j\right) P\left(\omega_1 \mid x^i\right) \\
& -P\left(\omega_1 \mid x^j\right) P\left(\omega_2 \mid x^i\right)
\end{aligned}
\end{equation}
The classifiability of the dataset is given by
\begin{equation}
L=\sum_i P(x^i) C\left(x^{(i)}\right),
\end{equation}\label{eqncf}
where $P(x^i)$ is the number of patterns in the neighbourhood of $x^{(i)}$ divided by the number of samples $N$. 
Given a regression dataset \textbf{R}, we first determine the equivalent cCassification dataset \textbf{C}. Regressability of \textbf{R} is determined as the classifiability of the equivalent classification dataset \textbf{C}. Figure 7 illustrates the concept on a few examples.

\begin{figure}[!t]
  \centering

  \subfloat[$z=x$]{%
    \includegraphics[width=0.18\linewidth]{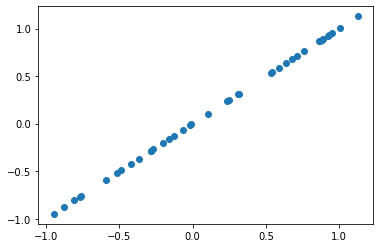}}%
  \hfill
  \subfloat[$z=x^2$]{%
    \includegraphics[width=0.18\linewidth]{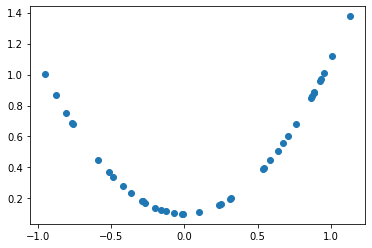}}%
  \hfill
  \subfloat[$z=x^3$]{%
    \includegraphics[width=0.18\linewidth]{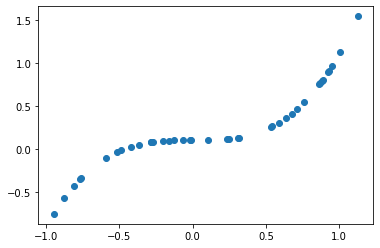}}%
  \hfill
  \subfloat[$z=\sin(x)$]{%
    \includegraphics[width=0.18\linewidth]{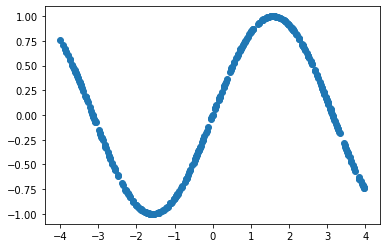}}%
  \hfill
  \subfloat[$z=\sin(x^2)$]{%
    \includegraphics[width=0.18\linewidth]{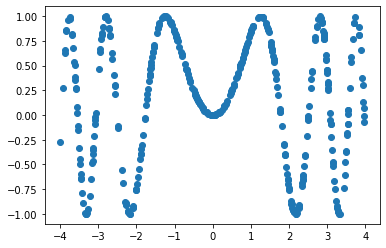}}%

  \par\medskip

  \subfloat[$z=x$]{%
    \includegraphics[width=0.18\linewidth]{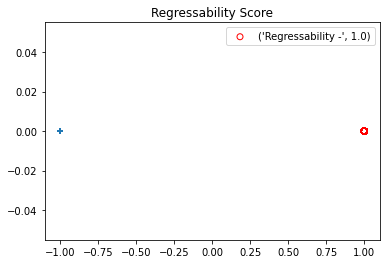}}%
  \hfill
  \subfloat[$z=x^2$]{%
    \includegraphics[width=0.18\linewidth]{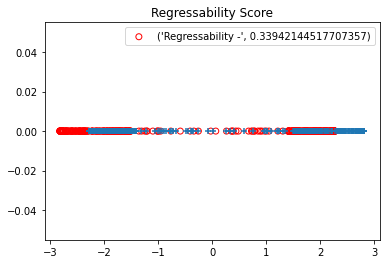}}%
  \hfill
  \subfloat[$z=x^3$]{%
    \includegraphics[width=0.18\linewidth]{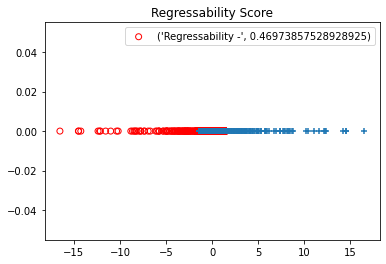}}%
  \hfill
  \subfloat[$z=\sin(x)$]{%
    \includegraphics[width=0.18\linewidth]{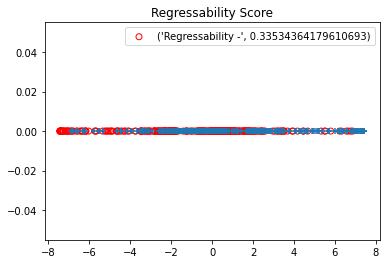}}%
  \hfill
  \subfloat[$z=\sin(x^2)$]{%
    \includegraphics[width=0.18\linewidth]{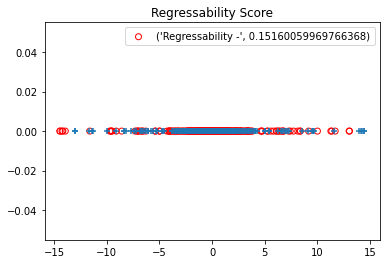}}%

  \caption{(a)--(e) Plots of selected 1-D functions, and (f)--(j) equivalent classification problems with computed regressability scores.}
  \label{fig:regressability}
\end{figure}

Figure 7 provides some insight into regressability. The linear function $z = x$ has a regressability score of 1.0; the equivalent classification samples are linearly separable. The regressability of $z = x^2$ is lower than that of $z = x^3$, because the monotonic slope of $z = x^3$ makes it easier to approximate with a line. This observation is with the caveat that trends may differ in the case of noisy data. Finally, $z = sin(x)$ has higher regressability than $z = sin(x^2)$. Table \ref{regtable1} includes a few more illustrative examples. Data samples in the table are taken from standard functions; the noise added is from a Gaussian distribution with mean 0 and variance 1.

\begin{table}[ht]
    \centering
    \caption{Regressability Examples}
    \begin{tabular}{lcc}
        \toprule
        Function & Regressability \\
        \midrule
        $Y = X $ + Noise & 0.99 \\
        $Y = X^2 $ + Noise & 0.34 \\
        $Y = X^3 $ + Noise & 0.31 \\
        $Y = X^5 $ + Noise & 0.18 \\
        $Y = (((X+1)^2)*(X-3)^2) $ + Noise & 0.26 \\
        $Y = (((X+3)^3)*((X-2)*(X+1)*(X+2))^2) $ + Noise & 0.28 \\
        $Y=(((X+3)^2)*((X+1)*(X+2))^2) $ + Noise & 0.42 \\
        $Y = \sin(X) $ + Noise & 0.43 \\
        $Y = \sin(X^2) $ + Noise & 0.15 \\
        $Y = X^2 + \sin(X^2) $ + Noise & 0.34 \\
        \bottomrule
    \end{tabular}
\end{table}\label{regtable1}
The availability of an equivalent classification problem enables a large repertoire of theoretical results and techniques traditionally used for classification, to be adapted for regression. For example, the notion of margin can be understood clearly in the context of equivalent samples. In the next section, we use the preceding discussion, to develop a different approach to regression using neural networks. We train a neural network to learn a linearizing map. Given a set of binary classification samples at the input of a neural network, image vectors at the output layer are such that samples of the same class are mapped close to each other in the output space, and those of different classes are as distant from each other as possible. The loss function tries to minimize the within-class scatter, and maximize the between-class scatter.

\section{Learning a Linearizing Map}\label{J4}
\begin{figure}[h]
\centering
\includegraphics[width=0.8\textwidth]{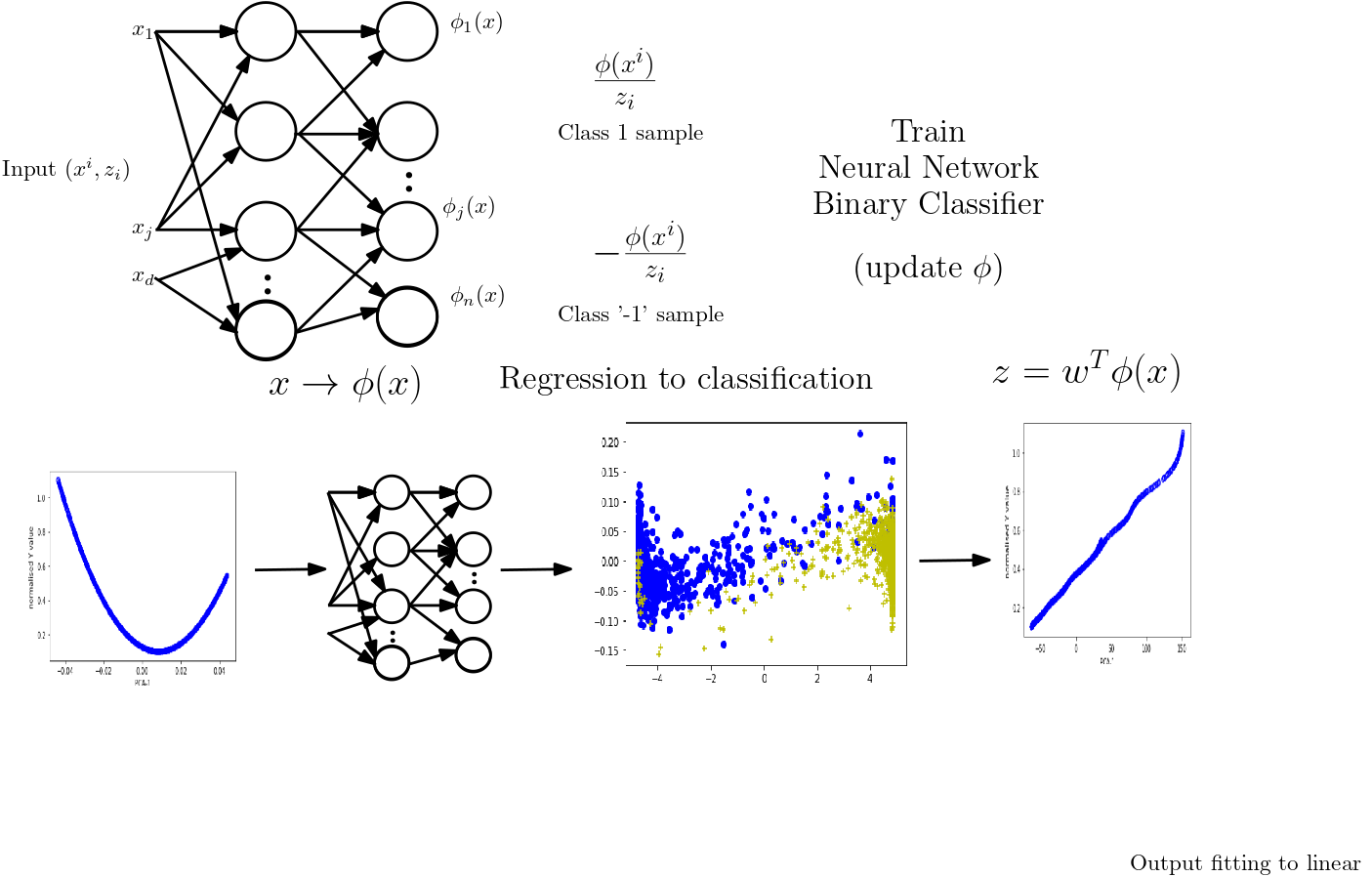}
  \caption{Learning a linearizing map $\phi(x)$ given samples $(x^i, z_i)$}
  \label{linmap1}
\end{figure}

\begin{figure}[!t]
  \centering
  \subfloat[$z = x^2 + \eta$]{%
    \fbox{\includegraphics[width=0.18\linewidth]{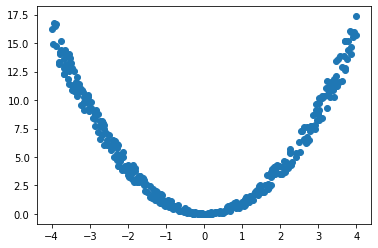}}}%
  \hfil
  \subfloat[Equivalent classification problem]{%
    \fbox{\includegraphics[width=0.18\linewidth]{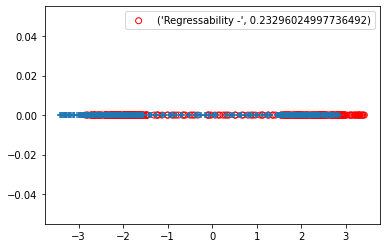}}}%
  \hfil
  \subfloat[after 100 epochs]{%
    \fbox{\includegraphics[width=0.18\linewidth]{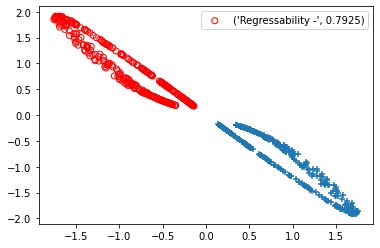}}}%
  \hfil
  \subfloat[after 500 epochs]{%
    \fbox{\includegraphics[width=0.18\linewidth]{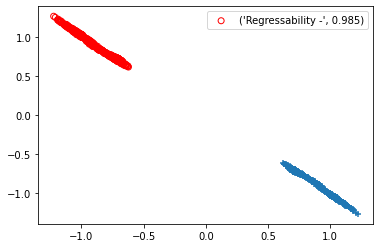}}}%
  \hfil
  \subfloat[after 1000 epochs]{%
    \fbox{\includegraphics[width=0.18\linewidth]{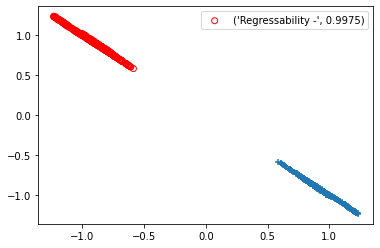}}}%

  \caption{Neural network training: (a) samples from $z=x^2+\text{noise}$; (b) samples of the equivalent classification problem; (c)--(e) projection of output-layer samples $\phi(x^i)$ onto the first principal component, with the vertical axis indicating the regression target ($z$), taken after 100, 500, and 1000 training epochs, respectively.}
  \label{linmapex1}
\end{figure}

Figure \ref{linmap1} illustrates the proposed approach.
Consider a neural network that maps input samples $x$ into a space $\phi(x)$ at the output layer. An input regression sample $(x^i, z_i)$ is mapped to its image $\phi(x^i)$ at the output layer. The equivalent binary classification samples at the output are $\frac{\phi(x^i)}{z_i}$ and $\frac{-\phi(x^i)}{z_i}$. The neural network loss function is designed to minimize the scatter within samples of the same class and maximize the distance between samples of different classes. Figure \ref{linmapex1}(a) shows the plot of the function $z = x^2$. Gaussian noise with zero mean and unit variance has been added to the target $z$. Figure \ref{linmapex1}(b) shows how the equivalent binary classification  samples are distributed; colours indicate class labels. Figure \ref{linmapex1}(c)-(e) show the output layer image samples i.e. $\phi(x^i)$, projected onto the first two principal components of the space of image samples. Note that the regressability of samples in the $\phi()$ space improves as the neural network gets trained over epochs.

\begin{figure}[!t]
  \centering

  \subfloat[$z=x^3+\eta$]{%
    \fbox{\includegraphics[width=0.18\linewidth]{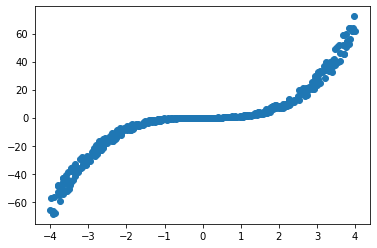}}}%
  \hfill
  \subfloat[Equivalent classification problem]{%
    \fbox{\includegraphics[width=0.18\linewidth]{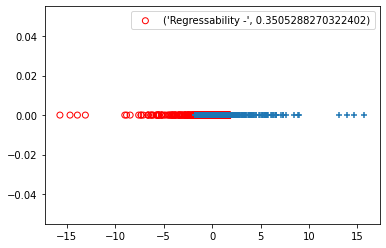}}}%

  \par\medskip

  \subfloat[after 100 epochs]{%
    \fbox{\includegraphics[width=0.18\linewidth]{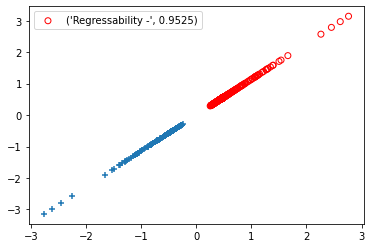}}}%
  \hfill
  \subfloat[after 500 epochs]{%
    \fbox{\includegraphics[width=0.18\linewidth]{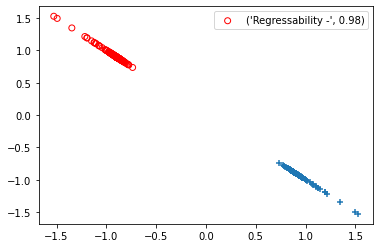}}}%
  \hfill
  \subfloat[after 1000 epochs]{%
    \fbox{\includegraphics[width=0.18\linewidth]{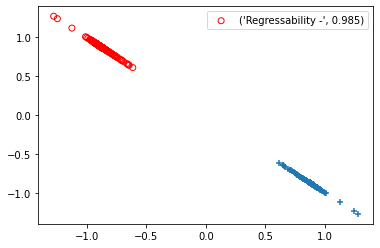}}}%

  \caption{Neural network training: (a) samples taken from $z=x^3+\text{noise}$; (b) samples of the equivalent classification problem; (c)--(e) projection of output-layer samples $\phi(x^i)$ onto the first principal component, with the vertical axis indicating the regression target ($z$), taken after 100, 500, and 1000 training epochs, respectively.}
  \label{fig:linmapex2}
\end{figure}

\begin{figure}[!t]
  \centering

  \subfloat[$z=\sin(x^2)+\eta$]{%
    \fbox{\includegraphics[width=0.18\linewidth]{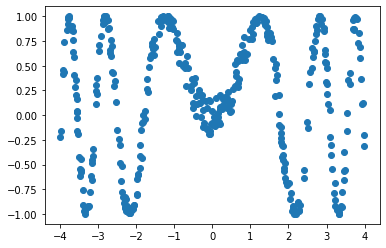}}}%
  \hfill
  \subfloat[Equivalent classification problem]{%
    \fbox{\includegraphics[width=0.18\linewidth]{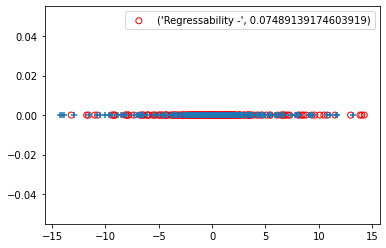}}}%

  \par\medskip

  \subfloat[after 100 epochs]{%
    \fbox{\includegraphics[width=0.18\linewidth]{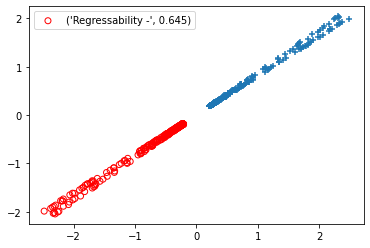}}}%
  \hfill
  \subfloat[after 500 epochs]{%
    \fbox{\includegraphics[width=0.18\linewidth]{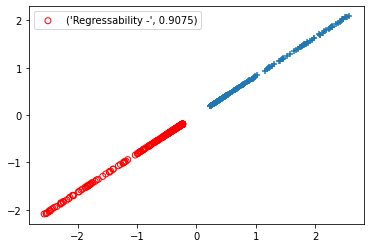}}}%
  \hfill
  \subfloat[after 1000 epochs]{%
    \fbox{\includegraphics[width=0.18\linewidth]{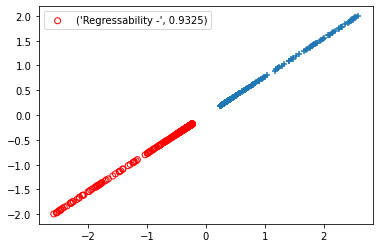}}}%

  \caption{Neural network training: (a) samples taken from $z=\sin(x^2)+\text{noise}$; (b) samples of the equivalent classification problem; (c)--(e) projection of output-layer samples $\phi(x^i)$ onto the first principal component, with the vertical axis indicating the regression target ($z$), taken after 100, 500, and 1000 training epochs, respectively.}
  \label{fig:linmapex3}
\end{figure}


Figures 10 and 11 illustrate the same with functions $z = x^3$ and $x = sin(x^2)$. One advantage of learning a linearizing map is better interpretability. The output layer features provide insight into a regressed function since it can be expressed as a linear combination of them.\\

Define
\begin{equation}
S_b = \left(\mu_1-\mu_2\right)\left(\mu_1-\mu_2\right)^T
\end{equation}\label{eqSb}
\begin{equation}
S_w =\frac{1}{2M} \sum_{i=1}^2 \sum_{j=1~,~ x^j \in Class~ i}^{M}\left(\phi(x_j -\mu_i\right)\left(\phi(x_j)-\mu_i\right)^T
\end{equation}\label{eqSw}
where $\mu_1$ and $\mu_2$ are the means of class 1 and class 2, respectively. The neural network uses the $J_4$ loss function proposed by Fukunaga \cite{fukunaga2013introduction}, that is given by
\begin{gather}
    J_4 = \frac{S_w}{S_b}
\end{gather}\label{J4}
Minimizing $J_4$ minimizes the within-class scatter and maximizes the between-class scatter. Algorithm \ref{alg:algorithm} summarizes the training procedure for a neural network using the $J_4$ loss function.

\begin{algorithm}[!h]
\caption{\textbf{$J_4$ Regression Algorithm}}
\label{alg:algorithm}
\textbf{Input}: Samples $(x^i, z_i), i = 1, 2, ..., M$\\
\textbf{Parameters}: ; Neural Network Architecture (no. of layers, neurons per layer), learning rate $\eta$.
\textbf{Output}: Feature map at the output layer, denoted by $\phi(x)$, where $x$ is the input to the first layer.
\begin{algorithmic}[1] 
\WHILE{not converged in epoch $i$}
\STATE Feed forward the sample from input space to output space $x \rightarrow \phi(x)$
\STATE Convert ($\phi(x^i), z_i)$), $i = 1,2,3..M$ into an equivalent classification problem
\STATE Class +1: $\frac{\phi(x^i)}{z_i}$
\STATE Class -1: $\frac{\phi(x^i)}{-z_i}$
\FOR{each class $c_i$}
\STATE calculate $\mu_i$: The sample mean for the $i$-th class 
\STATE Calculate the $J_4$ loss using (\ref{J4})
\STATE Update weights of all layers by back-propagating the gradient of the loss function.
\ENDFOR
\ENDWHILE
\STATE Denote the map learnt by the neural network as $\phi(x)$, where $x$ is the vector of inputs to the first layer of the neural network. Determine the co-efficients of a linear regressor $z = w^T\phi(x)$ to minimize the Mean Squared Error loss over all samples. 
\end{algorithmic}
\end{algorithm}

\begin{figure}[h]
    \centering
    \begin{minipage}{0.14\textwidth}
        \centering
        \includegraphics[width=\textwidth]{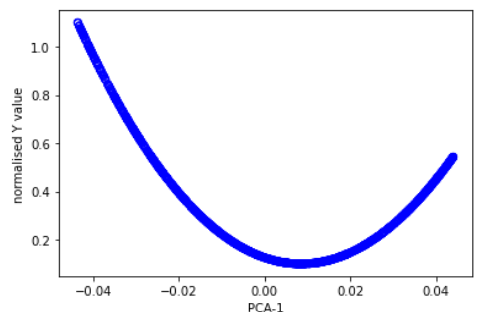}
    \end{minipage}
    \begin{minipage}{0.14\textwidth}
        \centering
        \includegraphics[width=\textwidth]{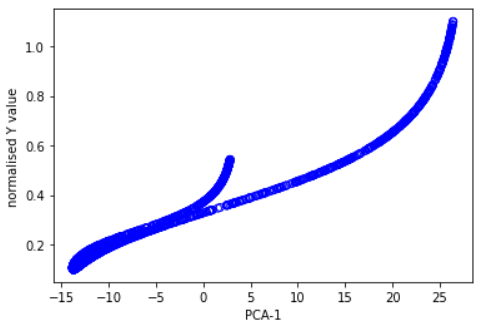}
    \end{minipage}
    \begin{minipage}{0.14\textwidth}
        \centering
        \includegraphics[width=\textwidth]{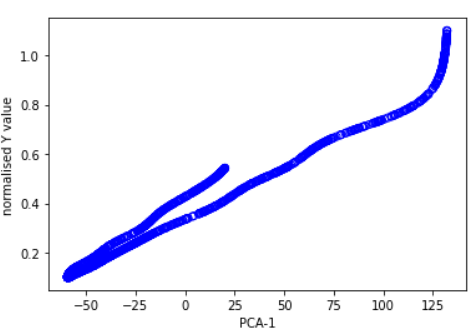}
    \end{minipage}
    \begin{minipage}{0.14\textwidth}
        \centering
        \includegraphics[width=\textwidth]{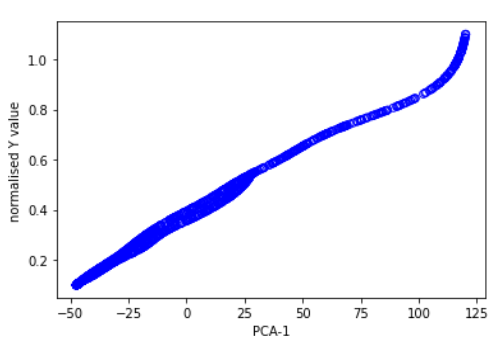}
    \end{minipage}
    \begin{minipage}{0.14\textwidth}
        \centering
        \includegraphics[width=\textwidth]{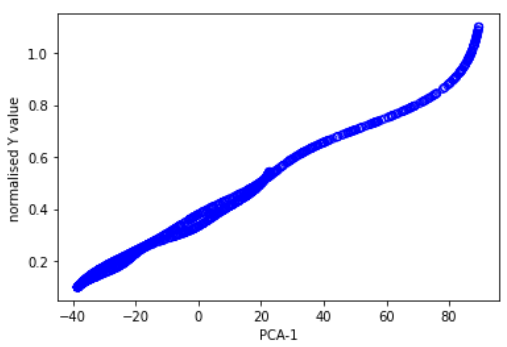}
    \end{minipage}
    \begin{minipage}{0.14\textwidth}
        \centering
        \includegraphics[width=\textwidth]{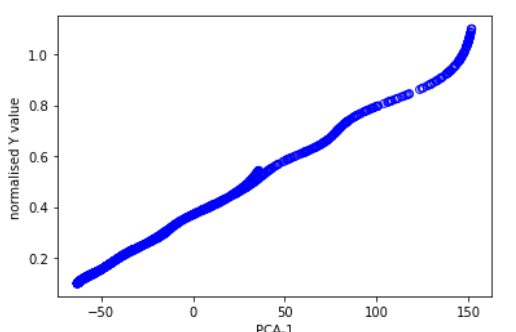}
    \end{minipage}

\caption{Samples of $z_i = (x^i)^2$ plotted vs. the projection of $\phi(x^i)$ onto the first principal component of output layer samples. The evolution indicates that as training progresses, the neural network learns a map so that the regressed output $z$ is a linear combination of neuron outputs in the final layer, i.e. $z = w^T \phi(x)$, where $w$ is a set of real co-efficients. Snapshots are at epoch 0, 2000, 4000, 6000, 8000, and 10000.}
    \label{linmapevol1}
\end{figure}

\begin{figure}[h]
    \centering
    \begin{minipage}{0.14\textwidth}
        \centering
        \includegraphics[width=\textwidth]{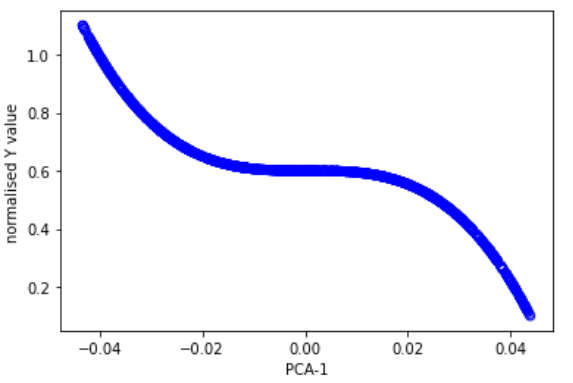}
    \end{minipage}
    \begin{minipage}{0.14\textwidth}
        \centering
        \includegraphics[width=\textwidth]{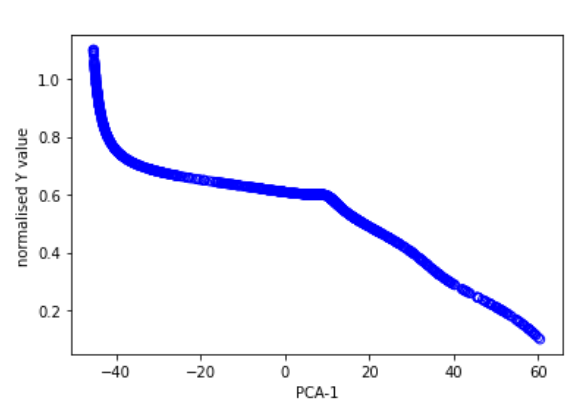}
    \end{minipage}
    \begin{minipage}{0.14\textwidth}
        \centering
        \includegraphics[width=\textwidth]{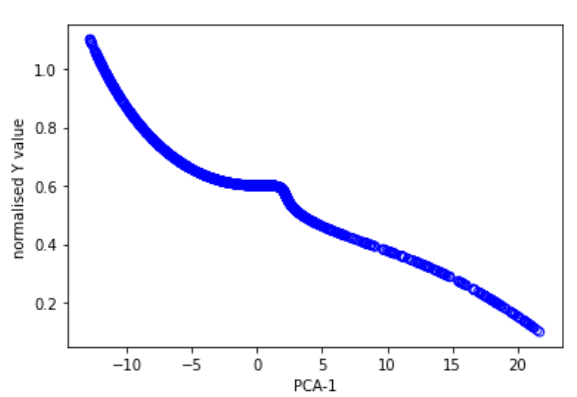}
    \end{minipage}
    \begin{minipage}{0.14\textwidth}
        \centering
        \includegraphics[width=\textwidth]{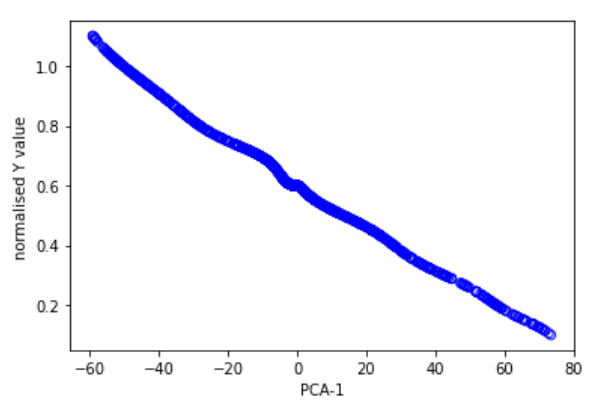}
    \end{minipage}
    \begin{minipage}{0.14\textwidth}
        \centering
        \includegraphics[width=\textwidth]{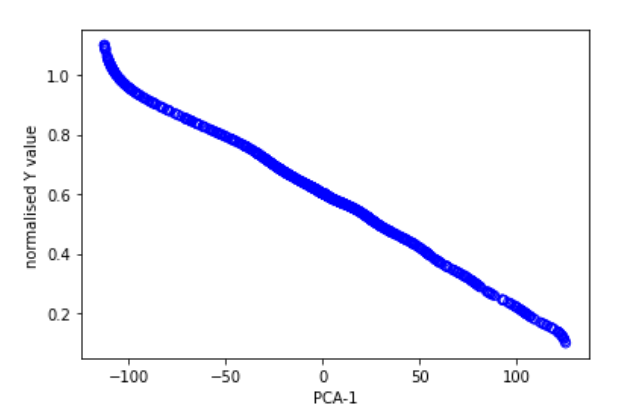}
    \end{minipage}
    \begin{minipage}{0.14\textwidth}
        \centering
        \includegraphics[width=\textwidth]{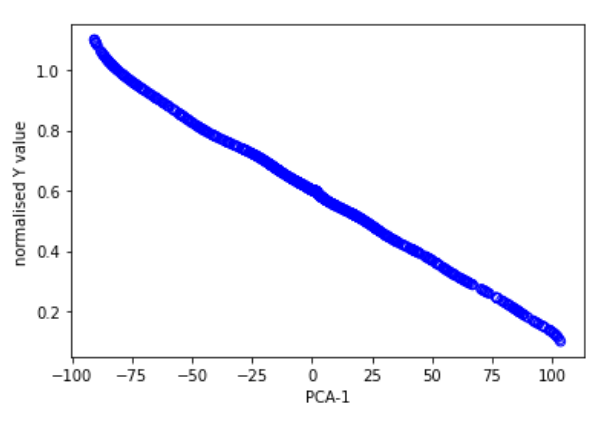}
    \end{minipage}

\caption{Samples of $z_i = (x^i)^3$ plotted vs. the projection of $\phi(x^i)$ onto the first principal component of output layer samples. The evolution indicates that as training progresses, the neural network learns a map so that the regressed output $z$ is a linear combination of neuron outputs in the final layer, i.e. $z = w^T \phi(x)$, where $w$ is a set of real co-efficients. Snapshots are at epoch 0, 2000, 4000, 6000, 8000, and 10000.}
    \label{linmapevol2}
\end{figure}

Figures \ref{linmapevol1} and \ref{linmapevol2} show how the map evolves when a neural network is trained with samples of $z = x^3$ and $z=x^2$, respectively. The network has 3  layers, with one input neuron, 5 hidden layer neurons, and 10 neurons in the output ($\phi$) layer. A tanh activation was used, and the learning rate was 0.01. The samples consisted of 1200 points randomly chosen in the domain (-2.0, 3.0).

Each subfigure shows $z_i$ plotted against the projection of $\phi(x^i)$ on the first principal component of output layer samples. The evolution indicates that as training progresses, the neural network learns a map so that the regressed output $z$ is a linear combination of neuron outputs in the final layer, i.e. $z = w^T \phi(x)$, where $w$ is a set of real co-efficients.

\section{Experimental Results}\label{results}
In order to provide a comprehensive comparison on the performance of the proposed approach, we refer to the very extensive experimental survey of regression methods by Delgado et al. \cite{fernandez2019extensive}. They consider 77 popular regression methods categorized into 19 families. These datasets were meticulously categorized into four distinct groups: "Small Easy," "Small Difficult," "Large Easy," and "Large Difficult." We chose datasets from the "Large Difficult" group, to facilitate comparisons with the most challenging and widely distributed examples. The training and testing methodology, including data folds, follows that used by \cite{fernandez2019extensive}. Table \ref{dataset_details} summarizes relevant statistics about the selected datasets.

\begin{table}[h]
    \centering
    \begin{tabular}{lll}
        \toprule
            \textbf{Dataset Name} & \textbf{No of Samples} & \textbf{Dimensions} \\
        \midrule
        1 Airfoil & 1,503 & 3 \\
        2 3DRoad & 4,34,874 & 4 \\
        3 Beijing pm25 & 41,758 & 12 \\
        4 Buzz Twitter & 583,250 & 77 \\
        5 Cuff less & 61,000 & 3 \\
        6 Facebook comment & 40,949 & 54 \\
        7 KEGG Relation & 54,413 & 22 \\
        8 Online news & 39,644 & 59 \\
        9 Greenhouse net & 955,167 & 15 \\
        10 Physico protein & 45,730 & 9 \\
        11 pm25 beijing dongsi & 24,237 & 13 \\
        12 pm25 beijing dongsi huan & 20,166 & 13 \\
        13 pm25 beijing nongzhanguan & 24,137 & 13 \\
        14 pm25 beijing us post & 49,579 & 13 \\
        15 pm25 chengdu shahepu & 23,142 & 13 \\
        16 pm25 chengdu caotangsi & 22,997 & 13 \\
        \bottomrule
    \end{tabular}
    \caption{Dataset Information}
    \label{dataset_details}
\end{table}

Table \ref{comparisons} summarizes comparisons of the proposed $J_4$ regression approach with results of the Average Neural Network (AvNNet) taken from \cite{fernandez2019extensive}. The AvNNet has been used as the baseline since it is the neural network based approach (see \cite{fernandez2019extensive}) in their survey.

\begin{table}[ht]
    \centering
    \caption{Test Data R2 Score for AvNN and J4 Regressor on Different Datasets}
    \begin{tabular}{lccc}
        \toprule
        \textbf{Sr No} & \textbf{Data Set} & \textbf{AvNN(Avg Neural Network)} & \textbf{J4 Regressor} \\
        \midrule
        1 & Airfoil & 0.892 & \textbf{0.902} \\
        2 & 3DRoad & 0.4681 & \textbf{0.9055} \\
        3 & Beijing\_pm25 & 0.6386 & \textbf{0.6531} \\
        4 & Buzz\_Twitter & 0.3929 & \textbf{0.9491} \\
        5 & Cuff\_less & 0.7242 & 0.6465 \\
        6 & Facebook\_comment & 0.8205 & 0.7167 \\
        7 & KEGG\_Relation & 0.9089 & \textbf{0.9483} \\
        8 & Online\_news & 0.1445 & -0.00012 \\
        9 & Greenhouse\_net & 0.1259 & \textbf{0.483} \\
        10 & Physico\_protein & 0.4728 & \textbf{0.5342} \\
        11 & pm25\_beijing\_dongsi & 0.6174 & \textbf{0.6436} \\
        12 & pm25\_beijing\_dongsihuan & 0.6495 & \textbf{0.6767} \\
        13 & pm25\_beijing\_nongzhanguan & 0.6312 & \textbf{0.6704} \\
        14 & pm25\_beijing\_us\_post & 0.6229 & \textbf{0.6445} \\
        15 & pm25\_chengdu\_shahepu & 0.4903 & \textbf{0.6042} \\
        16 & pm25\_chengdu\_caotangsi & 0.5121 & \textbf{0.5967} \\
        \bottomrule
    \end{tabular}
    \label{comparisons}
\end{table}

The tabulated results indicate that the imposition of a linearity constraint during the training of a Deep Neural Network led to an enhancement in model performance. Specifically, the R2 score achieved through the application of the $J_4$ regression method surpassed that of the Averaged Neural Network Model utilizing Mean Squared Error loss.

\section{Conclusion} \label{conclusion}

In this work, we proposed a novel approach to regression tasks by incorporating linearity constraints inspired by the mathematical principles governing straight lines. Our methodology focused on transforming regression problems into equivalent classification problems and enforcing linearity in the feature space. We utilized the $J_4$ loss function to achieve convergence of class points and maximize the separation between them.

Our approach demonstrated successful linearity in the feature space, allowing for the prediction of data points in both interpolated and extrapolated regions. We further extended our methodology to handle interpolation and extrapolation using Radial Basis Function Neural Networks, showcasing its effectiveness in capturing input data distribution and ensuring smooth predictions.

The experimental results on various datasets, including challenging "Large Difficult" datasets, revealed that our $J_4$ regression model outperformed the Averaged Neural Network Model with Mean Squared Error loss in terms of R2 scores. This suggests that enforcing linearity constraints during training can improve the performance of regression models, especially in scenarios with complex and widely distributed data.

In conclusion, our proposed regression methodology, with its focus on linearity and the $J_4$ loss function, presents a promising approach for addressing regression tasks. Further research and application of this methodology in diverse domains could provide valuable insights into its generalizability and effectiveness in UCI datasets.

\appendix
\section*{Appendix I}\label{Appendix1}
Consider a classification dataset given by $\{x^i, y_i\},~ i=$ 1, 2 ..., $M$, where $x^i \in \Re^n$, $y_i \in \{-1, 1\}$. We rewrite the SVC formulation (\ref{svcdualobj})-(\ref{svccons1}), where we assume that the separating hyperplane passes through the origin. 
 \begin{gather}
     \text{Minimize } \frac{1}{2}\|w\|^2 + C \sum_{i = 1}^M \left( q_i^+ + q_i^-\right) \label{svcobj}\\
     \text{subject to the constraints} \nonumber\\
     y_i(w^Tx^i) + \left( q_i^+ - q_i^-\right) = 1 \label{svccon1}\\
     q_i^+, q_i^- \geq 0~, \; i = 1, 2, ..., M \label{svccon2}\\
 \end{gather}
The Lagrangian for the above primal is given by
\begin{align}
     \mathcal{L}(w, q^+, q^-, \lambda, \eta^+, \eta^-) = \frac{1}{2}\|w\|^2 & + C \sum_{i = 1}^M \left( q_i^+ + q_i^-\right) - \sum_{i = 1}^M ~ \eta_i^+ q_i^+ -\sum_{i = 1}^M ~ \eta_i^- q_i^- \nonumber \\
     & +\sum_{i = 1}^M ~ \lambda_i \left[ 1 - y_i(w^Tx^i) - q_i^+ + q_i^- \right]  \label{lag0}\\
     -\infty \leq \lambda_i \leq \infty; \; \eta_i^+ \geq 0; \; \eta_i^- \geq 0
 \end{align}
The Karush-Kuhn-Tucker (K.K.T.) conditions are given by \cite{chandra2009numerical}
\begin{gather}
     \nabla_w \mathcal{L} = 0 \implies w - \sum_{i = 1}^M ~ \lambda_i y_i x^i = 0 \implies w = \sum_{i = 1}^M ~ \lambda_i y_i x^i \label{lag1}\\
     \frac{\partial \mathcal{L}}{q_i^+} = 0 \implies C - \eta_i^+ - \lambda_i = 0 \implies \eta_i^+ + \lambda_i = C \label{lag2}\\
     \frac{\partial \mathcal{L}}{q_i^-} = 0 \implies C - \eta_i^- + \lambda_i = 0 \implies \lambda_i - \eta_i^- = C \label{lag3}\\
    -\infty \leq \lambda_i \leq \infty  \label{lamlimit} \\
    \eta_i^+, \eta_i^- \geq 0 \label{etalimit}
\end{gather}
From (\ref{lag2}), (\ref{lag3}), (\ref{lamlimit}) and (\ref{etalimit}), we have
\begin{gather}
    \eta_i^+ = C - \lambda_i \geq 0 \implies \lambda_i \leq C \\
    \eta_i^- = C + \lambda_i \geq 0 \implies \lambda_i \geq -C
\end{gather}
The complementarity K.K.T. conditions are given by
\begin{gather}
    \lambda_i \left[ 1 - y_i(w^Tx^i) - q_i^+ + q_i^- \right] = 0\\
    \eta_i^+ q_i^+  = 0\\
    \eta_i^- q_i^-  = 0, ~ i = 1, 2, ..., ~M
\end{gather}

Substituting from (\ref{lag1})-(\ref{lag3}) into (\ref{lag0}), we obtain the dual as
\begin{gather}
    \textbf{Max}_{\lambda}~ -\frac{1}{2} \sum_{i = 1}^M ~ \sum_{j = 1}^M~ \lambda_i \lambda_j y_i y_j \left(x^i\right)^T x^j + \sum_{i = 1}^M~ \lambda_i \\
\textbf{subject to the constraints} \nonumber\\
-C \leq \lambda_i \leq C
\end{gather}
which may be rewritten as
\begin{gather}
    \textbf{Min}_{\lambda}~ \frac{1}{2} \sum_{i = 1}^M ~ \sum_{j = 1}^M~ \lambda_i \lambda_j y_i y_j \left(x^i\right)^T x^j - \sum_{i = 1}^M~ \lambda_i \\
\textbf{subject to the constraints} \nonumber\\
-C \leq \lambda_i \leq C
\end{gather}
Finally, we re-visit our assumption that the regressor is of the form $z = w^Tx$, i.e. the regressor passes through the origin and does not have an offset. Consider a regressor of the form $z = w^Tx + b$, where $b$ is the offset. We assume that at least one sample, say, $x^0$, lies on the line, i.e. $z_0 = w^Tx^0 + b$. Note that at this point we do not know $w$ or $b$, since these are determined after solving the optimization problem, and the equivalent classification samples need to be determined before solving for the classifier.\\
Sample $x^0$ is treated as a reference point. All samples $\left(x^i, z_i \right)$, $i = 1, 2, ...$, $M$ are replaced by $\left(x^i - x^0, z_i - z_0 \right)$. Note that $w^T(x - x^0)$ passes through the origin, and $b$ is not needed.\\
In principle, we could take any sample and treat it as the reference $x^0$. Of course, $x^0$ is excluded from training data. However, in practice, some samples may be noisy or outliers; choosing such samples is imprudent. We therefore choose
\begin{gather}
    x^0 = \frac{1}{M}~ \sum_{i = 1}^M~ x^i\\
    z_0 = \frac{1}{M}~ \sum_{i = 1}^M~ z_i
\end{gather}
Although this choice is not always optimal, it averages any noise present in sample locations $x^i$ or regression targets $z_i$.

\bibliographystyle{ieeetr}
\bibliography{reg}

\begin{thebibliography}{1}

\bibitem{torgo1996regression}
L.~Torgo and J.~Gama, ``Regression by classification,'' in {\em Advances in Artificial Intelligence: 13th Brazilian Symposium on Artificial Intelligence, SBIA'96 Curitiba, Brazil, October 23--25, 1996 Proceedings 13}, pp.~51--60, Springer, 1996.

\bibitem{salman2012regression}
R.~Salman and V.~Kecman, ``Regression as classification,'' in {\em 2012 Proceedings of IEEE Southeastcon}, pp.~1--6, IEEE, 2012.

\bibitem{pontil1998regression}
M.~Pontil, R.~Rifkin, and T.~Evgeniou, ``From regression to classification in support vector machines,'' 1998.

\bibitem{suykens1999least}
J.~A. Suykens and J.~Vandewalle, ``Least squares support vector machine classifiers,'' {\em Neural processing letters}, vol.~9, pp.~293--300, 1999.

\bibitem{fung2001proximal}
G.~Fung and O.~L. Mangasarian, ``Proximal support vector machine classifiers,'' in {\em Proceedings of the seventh ACM SIGKDD international conference on Knowledge discovery and data mining}, pp.~77--86, 2001.

\bibitem{dong2003feature}
M.~Dong and R.~Kothari, ``Feature subset selection using a new definition of classifiability,'' {\em Pattern Recognition Letters}, vol.~24, no.~9-10, pp.~1215--1225, 2003.

\bibitem{fukunaga2013introduction}
K.~Fukunaga, {\em Introduction to statistical pattern recognition}.
\newblock Elsevier, 2013.

\bibitem{fernandez2019extensive}
M.~Fern{\'a}ndez-Delgado, M.~S. Sirsat, E.~Cernadas, S.~Alawadi, S.~Barro, and M.~Febrero-Bande, ``An extensive experimental survey of regression methods,'' {\em Neural Networks}, vol.~111, pp.~11--34, 2019.

\bibitem{chandra2009numerical}
S.~Chandra, Jayadeva, and A.~Mehra, {\em Numerical optimization with applications}.
\newblock Alpha Science International, 2009.

\end{thebibliography}
\end{document}